%% file: main.tex
\documentclass[letterpaper, 10 pt, conference]{ieeeconf}
\IEEEoverridecommandlockouts
\usepackage{cite}
\usepackage{amsmath,amssymb,amsfonts,amsthm}
\usepackage{graphicx}
\usepackage{textcomp}
\usepackage{xcolor}
\usepackage{url}
\usepackage{tablefootnote}

\usepackage[ruled,vlined]{algorithm2e}
\usepackage{algorithmic}
\usepackage{subcaption}
\usepackage{multirow}

\input{notations.tex}

\title{\LARGE \bf
Risk Conditioned Neural Motion Planning
}

\author{Xin Huang$^{1}$, Meng Feng$^{1}$, Ashkan Jasour$^{1}$, Guy Rosman$^{2}$, and Brian Williams$^{1}$ 
 \thanks{$^{1}$Computer Science and Artificial Intelligence Laboratory (CSAIL), Massachusetts Institute of Technology, Cambridge, MA 01239, USA
 {\tt xhuang@csail.mit.edu }}
\thanks{$^{2}$Toyota Research Institute, Cambridge, MA 01239, USA}
\thanks{This article solely reflects the opinions and conclusions of its authors and not TRI or any other Toyota entity.}
}

\begin{document}

\maketitle

\begin{abstract}
Risk-bounded motion planning is an important yet difficult problem for safety-critical tasks. While existing mathematical programming methods offer theoretical guarantees in the context of constrained Markov decision processes, they either lack scalability in solving larger problems or produce conservative plans. Recent advances in deep reinforcement learning improve scalability by learning policy networks as function approximators.  
In this paper, we propose an extension of soft actor critic model to estimate the execution risk of a plan through a risk critic and produce risk-bounded policies efficiently by adding an extra risk term in the loss function of the policy network. We define the execution risk in an accurate form, as opposed to approximating it through a summation of immediate risks at each time step that leads to conservative plans. Our proposed model is conditioned on a continuous spectrum of risk bounds, allowing the user to adjust the risk-averse level of the agent on the fly.
Through a set of experiments, we show the advantage of our model in terms of both computational time and plan quality, compared to a state-of-the-art mathematical programming baseline, and validate its performance in more complicated scenarios, including nonlinear dynamics and larger state space.
\end{abstract}

\section{Introduction}
Motion planning is an important task in many robotics applications, such as rescue robots and autonomous vehicles. One of the most popular approaches for motion planning is reinforcement learning, which learns an optimal policy that minimizes the cost, through exploration and exploitation. Recently, deep reinforcement learning has been proposed to approximate the policy function or the cost function by deep neural networks, and has achieved great success in applications where the state and action space is high-dimensional, such as robotics manipulation \cite{rajeswaran2017learning} and Go game \cite{silver2016mastering}.

\begin{figure}
    \centering
    \includegraphics[width=0.48\textwidth]{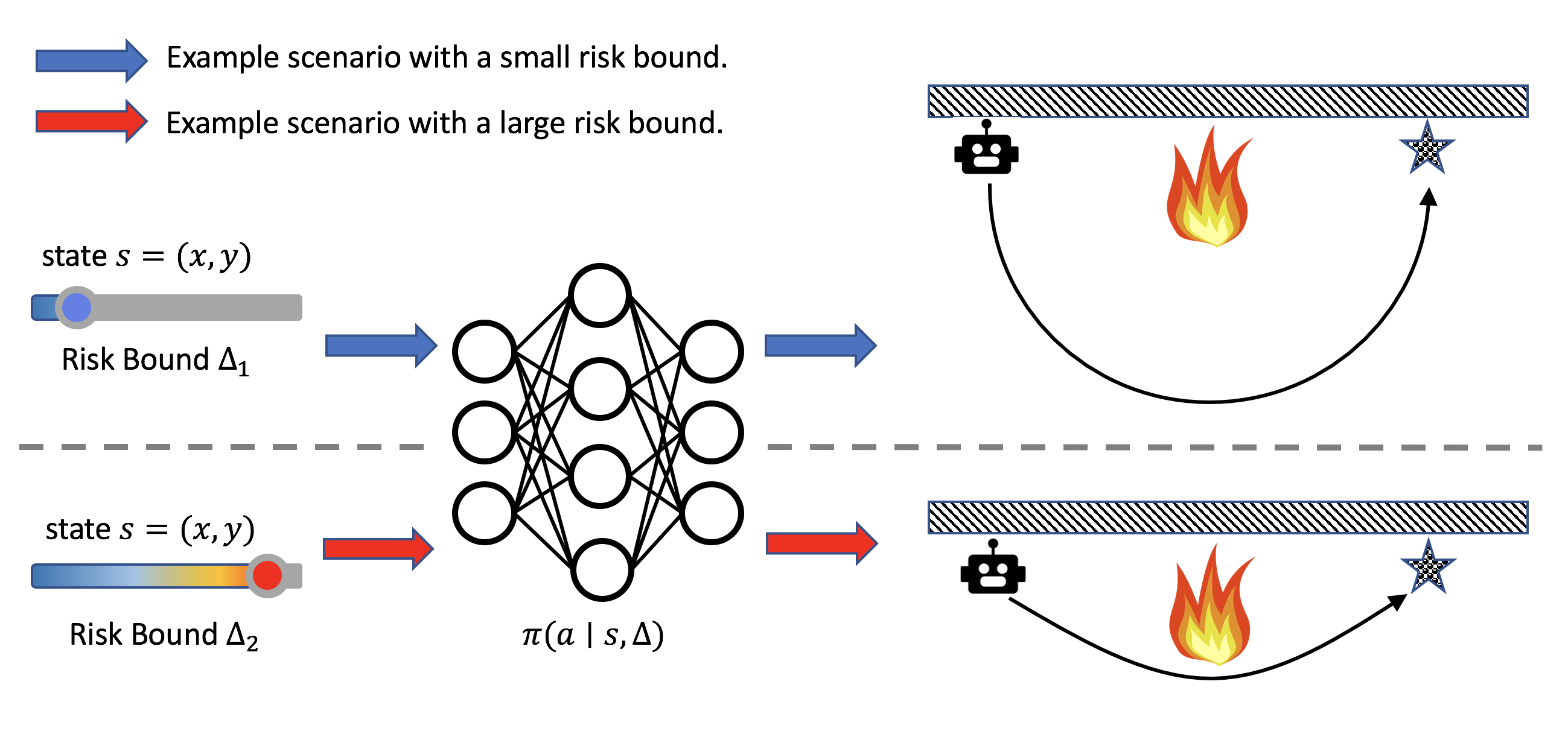}
    \caption{Overview of the proposed risk conditioned policy network. Given the current state and risk bound, the network generates an optimal action for the agent, subject to the risk bound. The policy is conditioned on an arbitrary upper bound on the probability of violating safety constraints, which has important applications in safety-critical domains such as rescue robots and autonomous vehicles. Top and bottom depict planning scenarios with a small risk bound and a large risk bound, respectively, using the same model.}
    \label{fig:model_diagram}
\end{figure}

Despite the success of reinforcement learning in the context of Markov decision processes (MDPs), its objective of naively minimizing a cost function is not sufficient in safety-critical tasks. For instance, in planning with obstacles for rescue robots (see Fig.~\ref{fig:model_diagram}) or autonomous vehicles, the objective is not only to minimize a certain cost function of the policy, such as time to a designated goal or fuel consumption, but also to satisfy the safety constraint of not colliding with obstacles, which may sacrifice the cost performance. 
Although in many cases, we can add the constraint as an infinite term in the cost function, this may lead to infeasible solutions, when the constraint is unavoidable (i.e. if we want to avoid an obstacle with unbounded uncertainties) and we can never guarantee safety. Therefore, it is more desirable to bound the probability of violating the constraint by a certain level, which is defined as a chance constraint in \cite{birge2011introduction}.

The requirement of satisfying additional constraint motivates the formulation of constrained Markov decision process (CMDP) \cite{altman1999constrained}, in which the agent needs to satisfy the constraint over an auxiliary cost while minimizing the cost function.
Although solving finite CMDP has been well studied with methods such as linear programming \cite{altman1999constrained} or mixed integer linear programming (MILP) \cite{ono2008efficient}, it remains a challenge to solve for high-dimensional or large CMDP instances. In cases where the constraint is probabilistic and the objective is to bound the probability of violating a constraint, \cite{thiebaux2016rao} introduces a special form of CMDP, named chance constraint MDP (CC-MDP), and proposes a heuristic forward search method in the discrete action space to find the desired policy efficiently, yet the performance heavily relies on the quality of heuristics.

Inspired by the recent success of deep reinforcement learning, we leverage deep neural networks to learn function approximators for the policy function and the risk (or chance constraint) function, similar to \cite{achiam2017constrained,pfeiffer2018reinforced,yu2019convergent}. 
The learned chance constraint is used in the policy optimization step to produce risk-bounded policies.
While existing methods assume a fixed upper bound on the chance constraint, our model is trained to generate policies conditioned on a range of continuous upper bound values. This allows us to control the agent with different risk tolerance levels on the fly, without the need to retrain the model. One motivating example is that in rescue tasks, as illustrated in Fig.~\ref{fig:model_diagram}, we may want to assign different risk bounds to the rescue robot based on the hard time limit, so that it can rescue the target in time. This requires the planner to generate policies in real-time conditioned on different risk bounds. 
Although it is possible to pre-generate a library of policies under different bound levels, it is less time efficient and space efficient, especially when the discretization of the bound requires a high resolution.
An overview of our approach is presented in Fig.~\ref{fig:model_diagram}.

Our contributions are as follows: i) We learn a risk estimator as a deep neural network to approximate the execution risk of a given policy, which provides task-specific information in addition to standard cost measures in MDP. ii) We leverage the learned risk estimator to learn a risk-bounded policy network, by adding the exceeded risk as an additional penalty in the policy loss, which balances performance and safety in the resulting plan. iii) Our policy network is conditioned on a continuous spectrum of upper risk bound, providing the flexibility to change the agent behavior on the fly at test time. iv) We provide a detailed comparison between a state-of-the-art MILP-based method and our proposed method, and show our method achieves better performances in terms of solution quality and computational time, without violating the risk bound.

\section{Related Work}
\subsection{Risk-Bounded Motion Planning}
\label{sec:rb_planning}
Risk-bounded motion planning problems have been studied in the forms of CMDPs and CC-MDPs in many works. Early methods show that CMDPs can be solved with linear programs (LPs) \cite{altman1999constrained,feinberg1996constrained} and MILPs \cite{dolgov2005stationary}. Most of these methods work with deterministic constraints. 
In \cite{ono2008efficient}, the constraint is modeled as the probability of failure, and an approach named Iterative Risk Allocation (IRA), is proposed to generate constrained policies by iteratively assigning the global probability constraint into individual probability budgets at each time step through a union bound, and solve for a MILP instance given the individual budgets. 
Aside from MILP-based methods, \cite{luders2010chance} proposes a sampling-based approach to generate constrained motion plans with probabilistic guarantees.
A common assumption of these works is that the global constraint can be expressed as the sum of sub-constraints over each time step, which may not hold if we define the constraint as the probability of collision over the entire path, since the joint disjunctive probability is not equivalent to the summation of probabilities of individual events. In fact, the summation is usually an upper bound of the joint probability, and optimizing for the summed constraints will likely lead to conservative policies. 

The conservatism is resolved in \cite{thiebaux2016rao}, which defines the risk as the probability of collisions in an exact form, and proposes RAO* to find optimal policies through heuristic forward search, which is demonstrated to work well in vehicle planning \cite{huang2018hybrid,huang2019online} and aircraft routing \cite{sungkweon2021anytime} domains. However, RAO* is limited to discrete action space due to its tree-based search approach.

Overall, due to the NP-hard nature of CMDP problems \cite{feinberg2000constrained}, both (MI)LP and search-based approaches suffer from scalability, especially in high-dimensional and continuous state space and action space. 

\subsection{Safe Reinforcement Learning}
Safety is a popular topic in reinforcement learning as many algorithms have been applied to real world applications, which requires the agent to explore the policy space while being safe to a certain extent. 
In this paper, we focus on RL methods that model risk as an explicit constraint in the overall objective function. In \cite{achiam2017constrained,chow2017risk}, the constrained RL problem is converted to an unconstrained RL problem using Lagrangian multiplier, and then solved by standard RL algorithms such as actor critic. Instead of solving the dual problem using the Lagrangian, \cite{yu2019convergent} approximates the objective function and the constraint function through quadratic surrogate functions, and solves for convex quadratically constrained quadratic program directly. As an extension to \cite{achiam2017constrained}, \cite{pfeiffer2018reinforced} combines a constrained policy optimization algorithm with imitation learning to bootstrap training time and achieves comparable performance. 
These methods define the constraint as a summed auxiliary cost from each time step, which leads to conservative policies if we model the constraint as the probability of collisions, as discussed in Sec.~\ref{sec:rb_planning}. Furthermore, the upper bound of the constraint is usually predefined as a fixed value, which would require the user to retrain the model if a different bound is needed. In our paper, we extend the state-of-the-art deep RL models to handle probabilistic constraints, in an accurate form, and flexible risk bounds, in the task of risk-bounded motion planning.


In addition to constrained MDP-based approaches, there exist works that improve safety by generating more samples in the risky region to bootstrap performance in critical scenarios \cite{andersson2017deep}; by using a safety layer at the end of a deep neural network to verify the safety of the resulting policy and replacing with a backup safe action if needed \cite{krasowski2020safe}; by proposing a reachability-based trajectory safe guard to ensure the safety of a policy \cite{shao2020reachability}, etc. In this paper, we focus on generating risk-bounded policies directly by modeling the risk as an explicit constraint in the objective function.

\subsection{Conditioned Reinforcement Learning}
Conditioned reinforcement learning provides great flexibility and improves generalizability when applying RL algorithms. A notable example is goal-conditioned RL \cite{levy2017learning,ghosh2018learning,nasiriany2019planning}, which produces goal-conditioned policies without assuming a fixed goal. This provides more capability to the agent as it can navigate to any goal locations in the environment without retraining. In our work, we are inspired by goal-conditioned policy and propose risk conditioned policy that generates policies conditioned on a continuous spectrum of upper risk bound.

\section{Problem Formulation}
\label{sec_problem}
In this work, we consider the risk-bounded motion planning problem in the absence of the agent models. More precisely, we aim at solving the chance constrained Markov decision process (CC-MDP) problem defined as follows:

\textbf{Definition 1. Chance-Constrained Markov Decision Process:} A Chance-Constrained Markov Decision Process (CC-MDP) is defined a tuple $\langle \mathcal{S}, \mathcal{A}, T, R, s_0, h, \mathcal{C}, \Delta\rangle$ as follows \cite{thiebaux2016rao}:
\begin{itemize}
\item $\mathcal{S}$ is a set of continuous states.
\item $\mathcal{A}$ is a set of actions for the agent model.
\item $T: \mathcal{S} \times \mathcal{A} \times \mathcal{S}  \to \mathbb{R}$ is a state transition function between the states.
\item $R: \mathcal{S} \times \mathcal{A} \to \mathbb{R}$ is a reward function.
\item $s_0$ is the initial state.
\item $h$ is the finite execution horizon.
\item $\mathcal{C}$ is a set of safety constraints defined over $\mathcal{S}$.
\item $\Delta$ is the upper risk bound.
\end{itemize} 

Given a definition of CC-MDP, the objective is to find a policy that maximizes the expected cumulative reward function:
\begin{equation} \label{reward_1}
    \pi^* = \arg \max_{\pi} \underset{( s_{t}, a_{t}) \sim \pi}{\mathbb{E}}\biggl[\sum_{t=0}^h R(s_t, a_t)\biggr],
\end{equation}
while satisfying the chance constraint with respect to the safety constraints $\mathcal{C}$:
\begin{equation} \label{er_1}
    er(s_0,\mathcal{C}| \pi) \leq \Delta,
\end{equation}
where $er$ is the execution risk defined as follows:
\begin{equation}
er^{\pi}(s_t) = er(s_t,\mathcal{C}|\pi) = 1 - \Pr \Bigg( \bigwedge_{i=t}^{h} Sa_i=1 \bigg|s_t,\pi \Bigg).  \label{eq:er}
\end{equation}
where $Sa_i$ is a Bernoulli random variable with value 1, when the state has not violated any constraint in $\mathcal{C}$, defined over $\mathcal{S}$, at time $i$. 

The execution risk can be computed in a recursive way as follows  \cite{thiebaux2016rao}:
\begin{equation}
    er^{\pi}(s_t) = r_b(s_t) + (1 - r_b(s_t)) \mathbb{E} [er^{\pi}(s_{t+1})],
    \label{eq:er}
\end{equation}
where $r_b(s_t)$ is the immediate risk at time $t$. An example of risk is the probability of collision between the agent and the obstacles. An important distinction between our work and existing CMDPs \cite{ono2008efficient,yu2019convergent} is that they approximate the execution risk as a sum of step-wise immediate risks:
\begin{equation}
    er^{\pi}_{approx}(s_t) = \sum_{i=t}^h r_b(s_i),
    \label{eq:er_a}
\end{equation}
which is an upper bound of $er^{\pi}(s_t)$, and solving for a policy following this definition leads to conservative results.

In this paper, we assume that the agent model and the reward function are hidden. Hence, we aim at solving the following planning problem: \\
\textbf{Risk-Bounded Motion Planning Problem:} Given a CC-MDP,
with unknown transition and reward functions, i.e., $T, R$, we look for a policy $\pi$ to maximize the expected cumulative reward function in \eqref{reward_1} with respect to the risk constraints in \eqref{er_1}.

\section{Approach}
In this section, we propose a deep neural network as a function approximator to solve the motion planning problem defined in Section \ref{sec_problem}. We first review the soft actor critic method, which is a popular reinforcement learning model that improves the stability of training compared to standard actor critic method. We then introduce our extended model, risk conditioned soft actor critic that includes a risk critic to estimate the probability of violating risk constraint, and show how we leverage the learned risk in policy optimization to learn risk-bounded policies. Finally, we present an algorithm to train our model through gradient descent.

\subsection{Soft Actor Critic}

Soft Actor Critic (SAC) \cite{haarnoja2018soft} is an off-policy actor critic deep reinforcement learning algorithm based on max entropy reinforcement learning. The objective is to find a policy that maximizes the maximum entropy:
\begin{equation}
\displaystyle \pi ^{*} =\underset{\pi }{\arg \max} \ \sum ^{h}_{t=0}\underset{( s_{t}, a_{t}) \sim \rho _{\pi }}{\mathbb{E}}[ R( s_{t} ,a_{t}) +\alpha \mathcal{H}( \pi ( \cdot |s_{t}))]
\label{eq:sac_obj}
\end{equation}

\noindent where temperature parameter $\alpha$ determines the relative importance of the entropy term versus the reward, and thus controls the stochasticity of the optimal policy \cite{haarnoja2018soft}; $\mathcal{H}( \pi ( \cdot |s_{t})$ is the entropy of the policy $\pi$ at state $s_t$; $\rho_\policy(s_t,a_t)$ denotes the state-action marginals of the trajectory distribution induced by a policy $\policy(a_t|s_t)$. 

As an actor critic algorithm, SAC alternates between policy evaluation, by estimating the value function for a policy, and policy improvement, by using the estimated value function to obtain a better policy.
More specifically, SAC leverages a state value function $V_{\psi}(s_t)$ parameterized by $\psi$, a soft Q-function $Q_{\theta}(s_t, a_t)$ parameterized by $\theta$, and a policy function $\pi_{\phi}(a_t|s_t)$ parameterized by $\phi$. These functions are modeled as deep neural networks (i.e. $V_{\psi}(s_t)$, $Q_{\theta}(s_t, a_t)$), or Gaussian parameters from a deep neural network (i.e. $\pi_{\phi}(a_t|s_t)$).

\subsubsection{Actor Model}
According to \cite{haarnoja2018soft}, the policy parameters can be learned by directly minimizing the following:
\begin{align}
J_\policy(\pparams) = \E{s_t\sim\mathcal{D}}{\log \policy_\pparams(f_\pparams(s_t)|s_t) - Q_\params(s_t, f_\pparams(s_t))},
\label{eq:reparam_objective}
\end{align}
where $\mathcal{D}$ is the replay buffer representing the distribution of previously sampled states and actions, $\pi_{\phi}$ is defined implicitly in terms of $f_\pparams( s_t)$, which is a neural network transformation that is used to reparameterize the policy, such that
\begin{align}
a_t = f_\pparams( s_t).
\end{align}
\subsubsection{Critic Model}
The soft Q-function parameters can be trained to minimize the soft Bellman residual:
\begin{equation}
J_Q(\theta) = \mathbb{E}_{(s_t, a_t)\sim \mathcal{D}}\biggl[{\frac{1}{2}\left(Q_{\theta}(s_t, a_t) - \hat Q(s_t, a_t)\right)^2}\biggr],
\label{eq:q_cost}
\end{equation}
with 
\begin{equation}
\hat Q(s_t, a_t) = R(s_t, a_t) + \gamma \mathbb{E}_{s_{t+1}\sim p}{V_{\psi}(s_{t+1})},
\end{equation}
where the soft value function $V_{\psi}$ is another deep neural network that is trained to approximate the value function. We refer to \cite{haarnoja2018soft} for more details of $V_{\psi}$.

\subsection{Risk Conditioned Soft Actor Critic}
\label{sec:rbsac}
In risk conditioned actor critic, we aim to learn an extra risk critic model $Q_{{er}_{\zeta}}$, parameterized by $\zeta$, that estimates the expected execution risk starting at a given state and acting according to the policy, defined in Eq.~\eqref{eq:er}.

\subsubsection{Risk Critic Model}
We use a risk critic model to estimate the execution risk, and it can be trained to minimize the L2 loss to the actual execution risk computed from data:
\begin{equation}
J_{Q_{er}}(\zeta) = \mathbb{E}_{(s_t, a_t)\sim \mathcal{D}}\biggl[{\frac{1}{2}\left(Q_{{er}_\zeta}(s_t, a_t) - \hat Q_{er}(s_t)\right)^2}\biggr],
\label{eq:q_cost}
\end{equation}
with 
\begin{equation}
\hat Q_{er}(s_t) = r_b(s_t) + (1 - r_b(s_t)) \mathbb{E}_{s_{t+1}\sim D}{\biggl[\hat Q_{er}(s_{t+1})\biggr]}.
\end{equation}
The gradient is computed as follows:
\begin{equation}
    \hat \nabla_{\zeta} J_{Q_{er}}(\zeta) = \nabla_{\zeta} Q_{{er}_\zeta}(s_t, a_t) (Q_{{er}_\zeta}(s_t, a_t) - \hat Q_{{er}}(s_t)).
\end{equation}

\subsubsection{Risk-Bounded Actor Model}
Given the risk critic model, we can update the loss function for the actor model as:
\begin{equation}
\begin{split}
J_\policy(\pparams) = & \mathbb{E}_{s_t\sim\mathcal{D}}\biggl[
\log \policy_{\pparams}(f_\pparams(s_t, \Delta)|s_t) - Q_{\phi}(s_t, f_\pparams(s_t, \Delta)) \\
& + \lambda_{er} \textit{ReLU}\Big(Q_{{er}_{\zeta}}\big(s_t, f_{\pparams}( s_t, \Delta)\big) - \Delta\Big) \biggr],
\end{split}
\label{eq:rsac_policy_loss}
\end{equation}
where $\textit{ReLU}(x)$ returns $x$ if $x \geq 0$ and 0 otherwise. This extra term in the policy loss adds an penalty if the estimated execution risk is larger than the risk bound, with a coefficient $\lambda_{er}$ to balance performance and safety. Compared to standard SAC, the transformation function $f_{\pparams}$ is updated to include $\Delta$ as its input so that the policy is conditioned on the upper risk bound:
\begin{align}
a_t = f_\pparams( s_t, \Delta).
\end{align}

\subsection{Algorithm}
\begin{algorithm}[tb]
\caption{Risk conditioned SAC algorithm.}
\label{alg:soft_actor_critic}
\begin{algorithmic}
\STATE \mbox{Initialize parameter vectors $\phi$, $\theta$, $\zeta$, $\Delta \sim \textit{U}[0, 1]$.}
\FOR{each iteration}
	\FOR{each environment step}
	\STATE $\at \sim \policy_\pparams(\at|\st, \Delta)$
	\STATE $\stp \sim p_T(\stp| \st, \at)$
	\STATE $\mathcal{D} \leftarrow \mathcal{D} \cup \left\{(\st, \at, R(\st, \at), r_b({\st}), \Delta, \stp)\right\}$
	\ENDFOR
	\FOR{each gradient step}
	\STATE $\theta \leftarrow \theta - \lambda_Q \hat \nabla_{\theta} J_\Q(\theta)$
	\STATE $\zeta \leftarrow \zeta - \lambda_{Q_{er}} \hat \nabla_{\zeta} J_{Q_{er}}(\zeta)$
	\STATE $\pparams \leftarrow \pparams - \lambda_\policy \hat \nabla_\pparams J_\policy(\pparams)$
	\ENDFOR
\ENDFOR
\end{algorithmic}
\label{algo:rbac}
\end{algorithm}
The algorithm to train a risk conditioned soft actor critic is illustrated in Alg.~\ref{algo:rbac}.
The key difference to the standard soft actor critic algorithm \cite{haarnoja2018soft} is that we need to train a separate \textit{risk critic} that estimates the execution risk.
This requires us to collect the immediate risk $r_b$ at each state, so that we can use it as a supervisory cue to train the risk critic network.
In addition, we generate random upper risk bound samples $\Delta$ from a uniform distribution, so that our policy network can learn to produce risk-bounded policies conditioned on different bounds. This, in practice, allows the online adjustment of the upper risk bound $\Delta$, subsequently changing the aggressiveness of the agent's actions. 
In the gradient descent step, we compute the gradient of each network, and we refer to \cite{haarnoja2018soft} for the detailed derivation of the gradients for the policy network and soft $Q$-function network.


\section{Results}
In this section, we introduce two sets of experiments to validate our proposed model. The first set of experiments provide a detailed study comparing our method and a state-of-the-art MILP-based baseline in a maze environment that simulates robotics rescue tasks; the second set of experiments showcase the performance of our method in more complicated scenarios, including nonlinear dynamics and larger state space, which usually pose great challenges for MILP-based methods. 
In each experiment, we define the problem setup and introduce model details, followed by results and discussions.

\subsection{Maze Environment}
\subsubsection{Problem Setup}
In order to show that our model provides risk conditioned policies efficiently, we introduce a \textit{OneObstacle} maze environment that includes an obstacle between the start location and the goal location, as illustrated in Fig.~\ref{fig:maze_paths}. The agent state space $S \in \mathbb{R}^2$ represents the continuous position $(x, y)$ bounded by the maze boundary $[0, 10] \times [0, 10]$, and the action space $A \in \mathbb{R}^2$ represents the linear velocity $(v_x, v_y)$ with the magnitude bounded by 1. The dynamics of the agent follows linear change of states:
\begin{gather}
    \dot{x} = v_x,\quad \dot{y} = v_y. 
    \label{eq:linear_dynamics}    
\end{gather}

We consider an additive Gaussian noise to the agent state with 0 mean and standard deviation of 1 along each axis, to model the uncertainty. The unbounded uncertainty prevents us from finding a policy that is risk-free; instead, we can only find a policy that bounds the probability of violating the risk constraint, in which the risk is defined as the probability of the agent colliding with any obstacles. While there exist a number of efficient risk assessment methods \cite{schmerling2016evaluating,wang2020fast}, we use a naive Monte Carlo sampling method to compute the collision risk for its simplicity.

\begin{figure*}[t!]
\vspace{3mm}
\begin{minipage}{0.33\textwidth}
\begin{subfigure}[b]{0.8\columnwidth}
    \includegraphics[width=6.6cm]{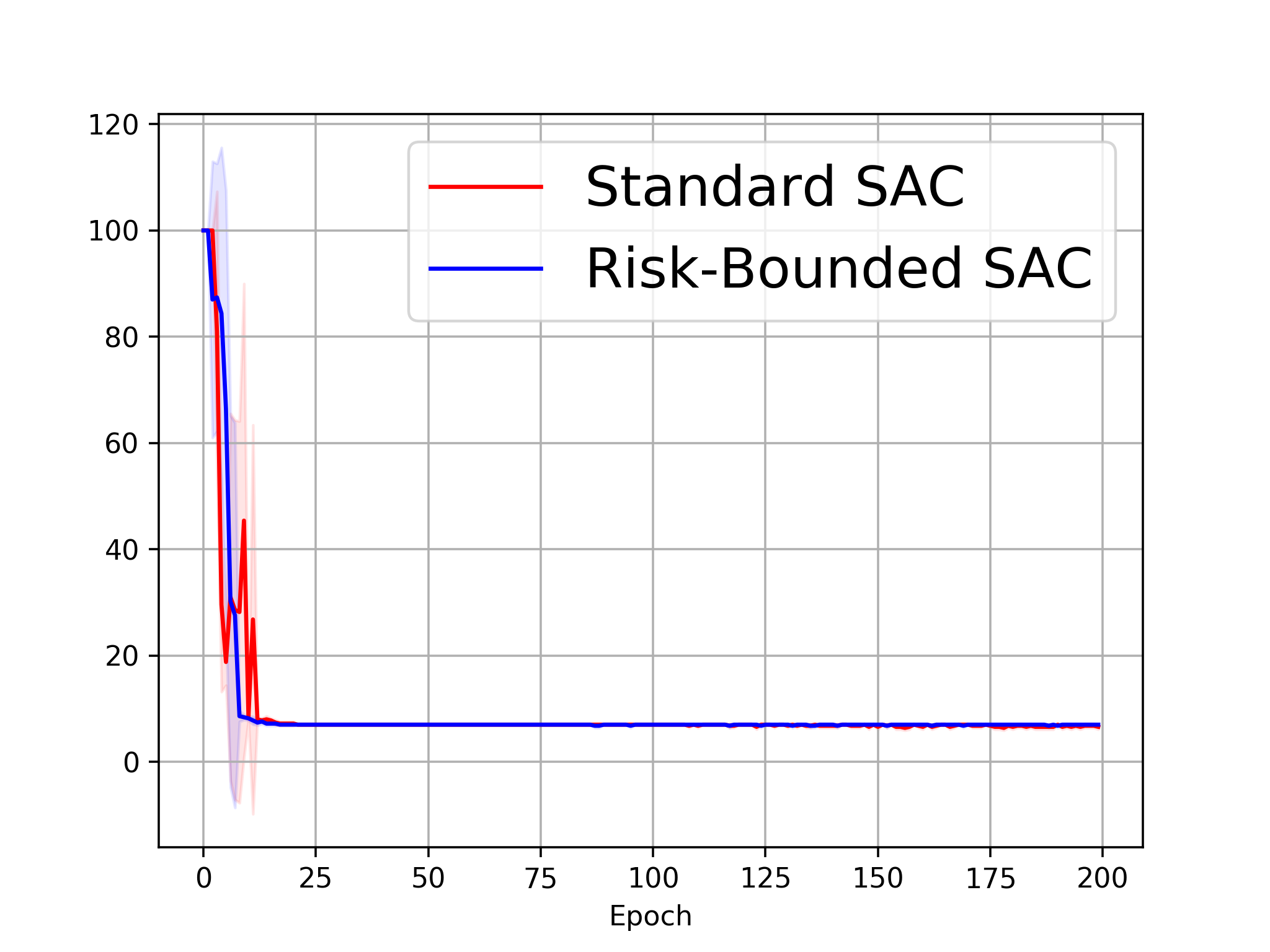}
  \end{subfigure}
  \begin{center}
  (a) Number of steps to goal.
  \end{center}
\end{minipage}
\begin{minipage}{0.33\textwidth}
\begin{subfigure}[b]{0.8\columnwidth}
     \includegraphics[width=6.6cm]{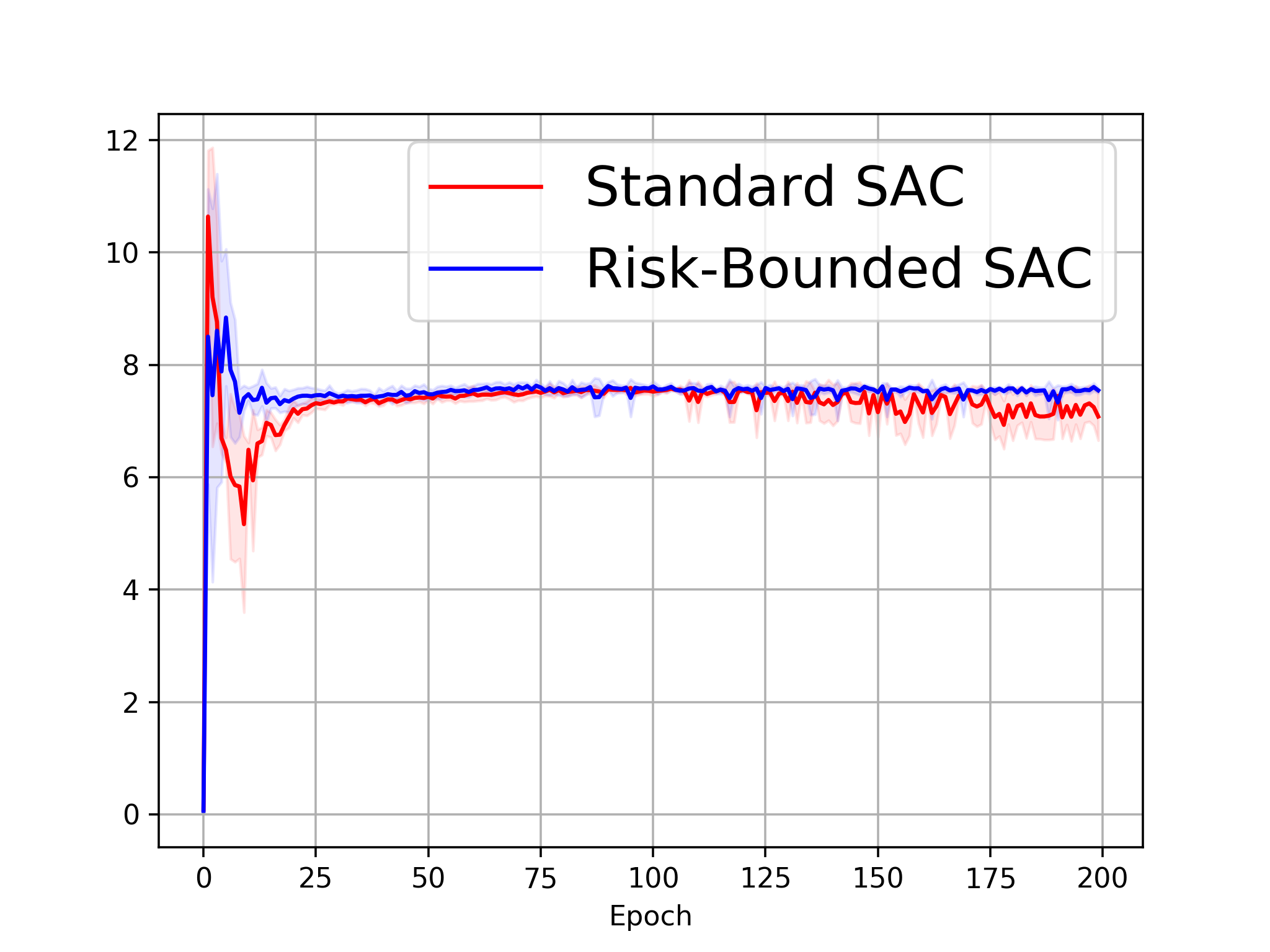}
  \end{subfigure}
  \begin{center}
  (b) Distance traveled to goal.
  \end{center}
\end{minipage}
\begin{minipage}{0.33\textwidth}
\begin{subfigure}[b]{0.8\columnwidth}
     \includegraphics[width=6.6cm]{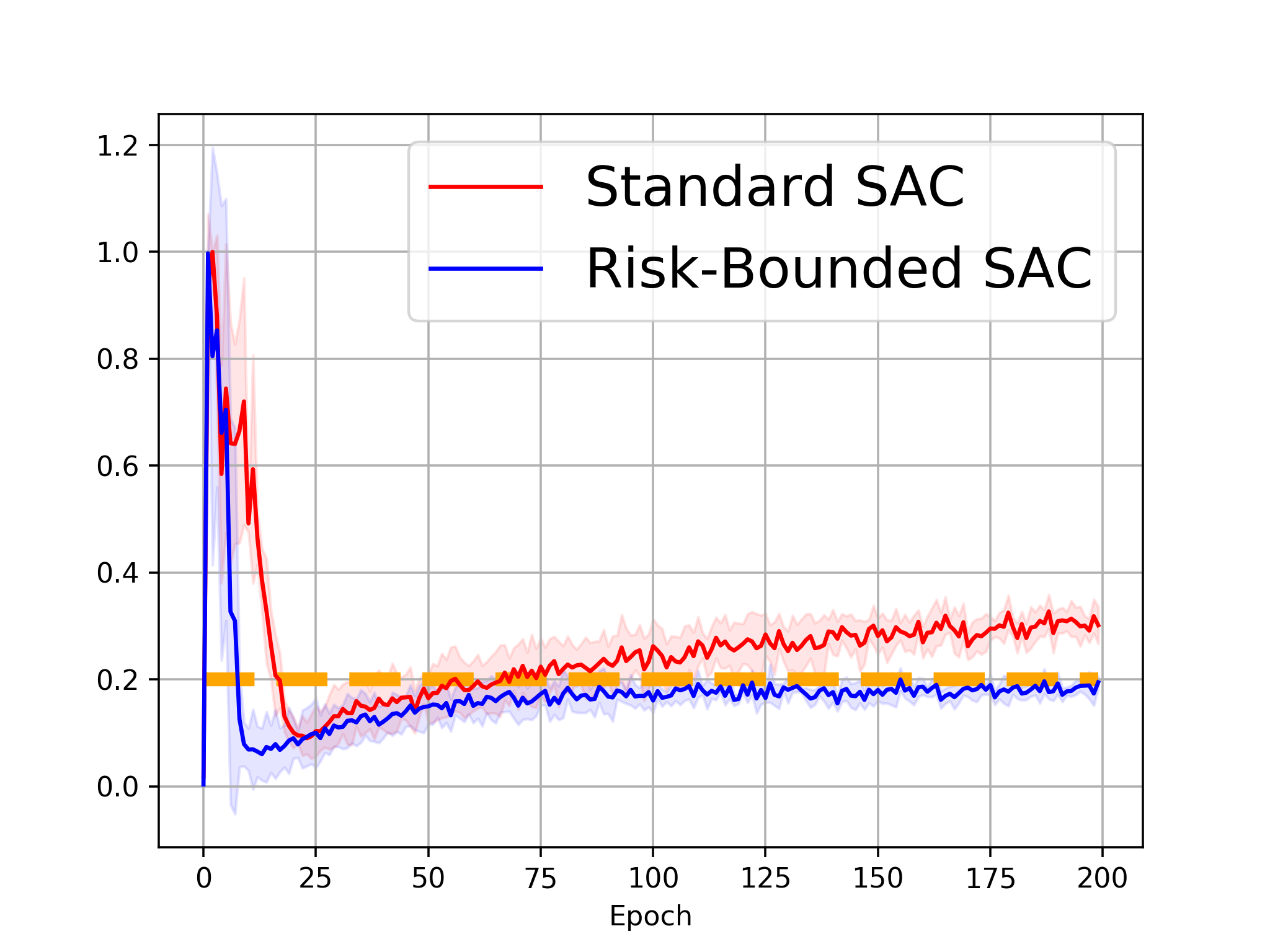}
  \end{subfigure}
  \begin{center}
  (c) Execution risk.
  \end{center}
\end{minipage}
  \caption{Evolution of evaluation metrics over runs from 5 randoms seeds, between standard SAC and our model, given a fixed upper risk bound $\Delta = 0.2$ in \textit{OneObstacle} maze environment. Shaded area represents one standard deviation. (a) Our model converges to the same number of time steps as SAC. (b) Our model converges to a distance slightly larger than SAC, due to additional risk constraints. (c) Our model converges to risk level below the risk bound, as visualized by the orange line.}
  \label{fig:sac_comparisons}
\end{figure*}

\subsubsection{Model Details}
\label{sec:simple_maze}
To find risk conditioned policies, we learn a risk conditioned SAC model\footnote{The source code will be open sourced in the near future at \url{https://github.com/cyrushx/risk_sac}.} described in Sec.~\ref{sec:rbsac}. The model includes two $Q$ function estimators as 3-layer multilayer perceptrons (MLPs) with hidden dimensions of 256, and two target $Q$ function estimators as 3-layer MLPs with the same structure. The policy network is composed of 3-layer MLPs with hidden dimensions of 256 and outputs a mean vector and a log standard deviation vector, which are further applied with an invertible squashing function (\textit{tanh}) to generate the bounded distribution for the resulting action.
In addition, our model includes a risk estimator with the same structure as the $Q$ function estimators. 

Our model is implemented in PyTorch, based on the open-source rlkit library\footnote{Source code at \url{https://github.com/vitchyr/rlkit}.}. At training time, we train for 200 epochs with a learning rate of 3e-4, a batch size of 256, and a reward discount factor of 0.99. The coefficient of risk penalty term $\lambda_{er}$ is 10. At test time, we evaluate the path generated from the learned policy and obtain the evaluation metrics, including distance traveled to goal and execution risk. The execution risk is computed using Monte Carlo sampling methods with 500 samples. We train and test our model in the same environment, as customary in many reinforcement learning tasks despite being prune to overfitting~\cite{cobbe2019quantifying}.

\subsubsection{Risk-Bounded Policy with a Fixed Upper Risk Bound} We start by validating the performance of our model given a fixed risk bound $\Delta = 0.2$ as input during training. This bound is selected arbitrarily -- in practice, it can be specified based on risk-averse level or requirement for plan efficiency. In addition, we train a standard soft actor critic (SAC) method that naively minimizes the cost function without reasoning about risk.

The comparisons between our model and standard SAC over 5 random seeds are illustrated in Fig.~\ref{fig:sac_comparisons}. We note that our model has been trained successfully to generate policies subject to the risk bound, as visualized by the orange line in Fig.~\ref{fig:sac_comparisons}(c). At the same time, it produces the policy with the same number of time steps and slightly worse traveled distance, compared to the standard SAC model. The worse traveled distance is due to the trade-off between performance and safety.

\subsubsection{Risk Conditioned Policy}
\begin{figure}[t!]
    \centering
    \includegraphics[width=0.41\textwidth]{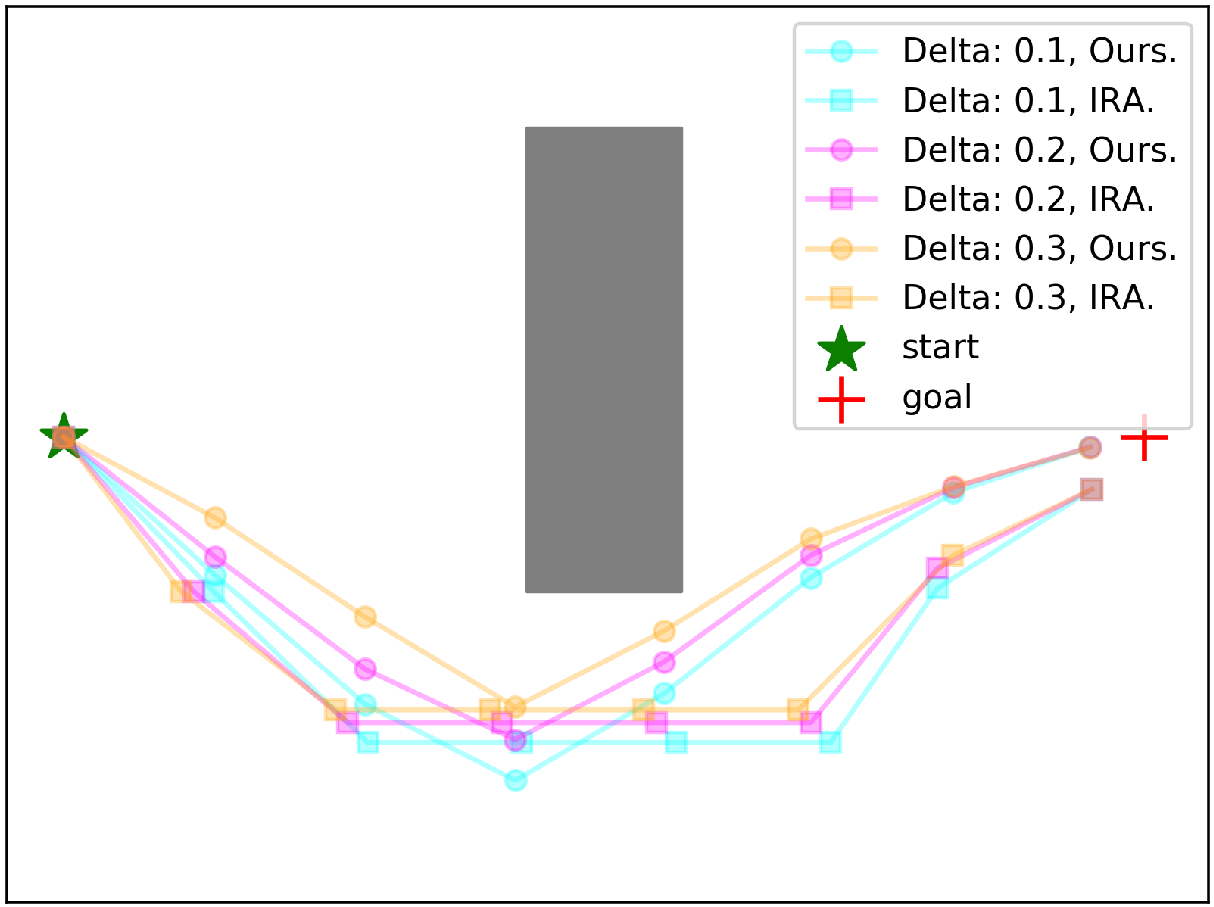}
    \caption{Visualization of paths generated by our model (circles) and IRA (squares), under 3 risk bounds represented in different colors, in \textit{OneObstacle} maze.}
    \label{fig:maze_paths}
\end{figure}

Next, we train our model by providing random risk bounds in the data so that it can learn risk-bounded policies conditioned on different upper risk bound levels. At test time, we visualize the planned trajectories by varying values for $\Delta$, as shown in Fig.~\ref{fig:maze_paths}. In the same figure, we visualize the risk-bounded policies generated by a state-of-the-art MILP-based solver, IRA, given the same upper risk bound levels and environment configurations. When the bounds become smaller, both models generate paths that are further away from the obstacle to improve safety. Both our model and IRA are run on the same hardware without GPU to make a fair comparison.

In Table~\ref{tab:path_comparison}, we compare the distance traveled to goal and computational time between our method and IRA. The results show that on average, our method is able to reduce the computational time by 93.87\%, demonstrating great time efficiency by using deep neural networks. Furthermore, it improves the plan quality, in terms of distance traveled to goal, by approximately 3.56\%, compared to IRA that produces conservative plans by assigning the overall risk constraint into each time step through a union bound. The table also shows the actual execution risk for each plan, which verifies that our plans are bounded by the desired $\Delta$ values. Overall, our method successfully produces risk-bounded policies efficiently without sacrificing the plan quality.

\begin{table}[]
\centering
\begin{tabular}{c|ccc|ccc}
\hline
\multirow{2}{*}{$\Delta$} & \multicolumn{3}{c|}{IRA}              & \multicolumn{3}{c}{Ours}            \\ \cline{2-7} 
                          & Distance{[}m{]} & Time{[}s{]} & Risk & Distance{[}m{]} & Time{[}s{]} & Risk \\ \hline
0.1                       & 8.20           & 0.16       & 0.024    & 8.04            & 0.0128      & 0.055    \\
0.2                       & 8.04           & 0.15       & 0.051    & 7.75            & 0.0085      & 0.125    \\
0.3                       & 7.93           & 0.16       & 0.071    & 7.52            & 0.0075      & 0.242    \\ \hline
\end{tabular}
\caption{Comparison between our model and IRA in \textit{OneObstacle} maze. Our model achieves better plan quality and computational time, subject to the risk bound.}
\label{tab:path_comparison}
\end{table}

\subsection{Dubins Car Model}
\subsubsection{Problem Setup}
In this experiment, we want to demonstrate the performance of our model with nonlinear dynamics that are usually difficult to be modeled by a MILP, which is limited to linear systems. More specifically, we consider the Dubins car model \cite {dubins1957curves} that is commonly used for real world ground robotics platforms \cite{giordano2009shortest}:
\begin{gather}
    \dot{x} = v\cos(\theta), \quad \dot{y} = v\sin(\theta), \\
    \dot{\theta} = u_{\theta}, \quad \dot{v} = u_{v}.
    \label{eq:dubins_path}
\end{gather}

The test environment is a standard \textit{TwoRooms} maze \cite{eysenbach2019search} that is composed of two rooms connected by two paths, as visualized in Fig.~\ref{fig:dubins}.

\subsubsection{Model Details}
We train a model with the same structure and parameters as in Sec.~\ref{sec:simple_maze}, except the output from the policy network changes from velocities to angular velocities and accelerations for Dubins car model. The model is trained for 500 epochs with a risk penalty term of 20, due to the nonlinear dynamics. Despite the longer convergence time, our system achieves similar run-time complexity, by leveraging a model of the same size.

\subsubsection{Risk Conditioned Policy}
\begin{figure}[t!]
    \centering
    \includegraphics[width=0.41\textwidth]{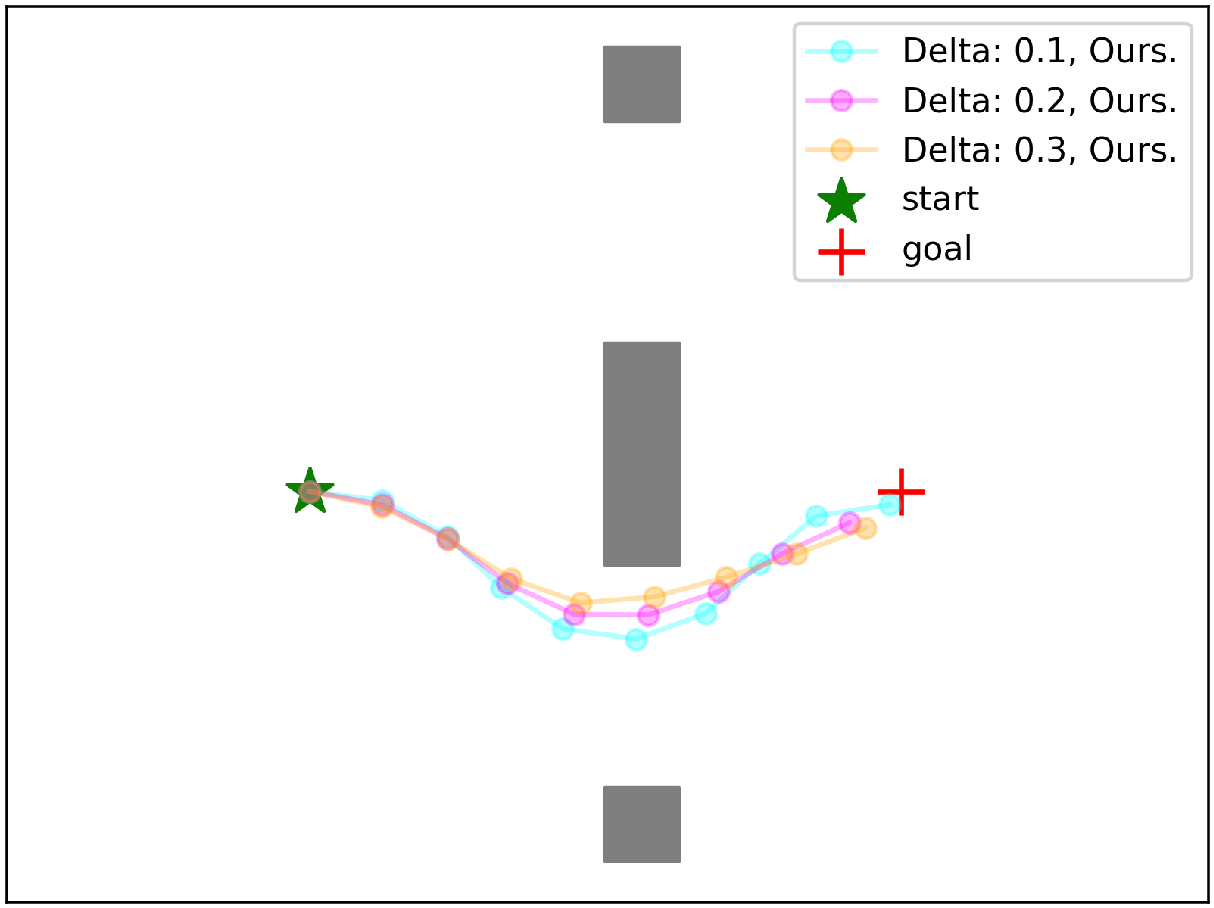}
    \caption{Visualization of paths generated by our model with Dubins car model, under different risk bounds, in \textit{TwoRooms}.}
    \label{fig:dubins}
\end{figure}

While it is usually infeasible to handle nonlinear dynamics in linear programs, we show that our model is capable of finding risk-bounded policies when the dynamics are nonlinear, as visualized in Fig.~\ref{fig:dubins}. The metrics are summarized in Table~\ref{tab:dubins}. We notice that under different risk bounds (i.e. 0.2 and 0.3), the risk-bounded plans have similar distances, which can be explained by the fact that the distance does not strictly decrease when the risk bound increases, due to the nonlinearity of the dynamics model.

\begin{figure}[t!]
    \centering
    \includegraphics[width=0.41\textwidth]{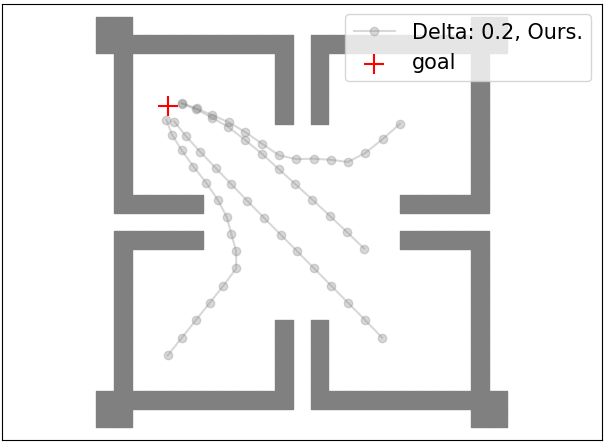}
    \caption{Visualization of paths generated by our model in \textit{FlyTrapBig} maze starting at different locations, under the same risk bound of 0.2. Our model successfully generalizes to random start locations in a large maze.}
    \label{fig:flytrapbig}
\end{figure}

\begin{table}[t!]
\centering
\begin{tabular}{c|ccc}
\hline
$\Delta$ & Distance{[}m{]} & Time{[}s{]} & Risk  \\ \hline
0.1     & 9.00            & 0.0094      & 0.076 \\
0.2     & 8.00            & 0.0071      & 0.142 \\
0.3     & 8.00            & 0.0072      & 0.214 \\ \hline
\end{tabular}
\caption{Dubins car model results in \textit{TwoRooms}. Our model produces risk-bounded plans under different risk bounds.}
\label{tab:dubins}
\end{table}


\subsection{\textit{FlyTrapBig} Maze}
\subsubsection{Problem Setup}
In this experiment, we validate our model in a larger maze, named \textit{FlyTrapBig} \cite{eysenbach2019search}, which has four times larger state size compared to the \textit{OneObstacle} maze. This requires more exploration in the space, and more importantly, requires more time steps to get to the goal, which poses great difficulty to the existing constrained MDP methods \cite{achiam2017constrained,yu2019convergent}. This is because these methods need to allocate the global risk bound into many individual time steps, which could end up returning infeasible solutions. On the other hand, our model avoids conservatism by using an accurate definition of risk and generates feasible solutions even when the number of steps to the goal is large.

\subsubsection{Model Details}
We train a model with the same structure and parameters as in Sec.~\ref{sec:simple_maze}, except with a larger epoch size of 1200 and a larger risk penalty coefficient $\lambda_{er}$ of 20, due to the larger state space and increased magnitude of plan rewards, respectively.

\subsubsection{Risk Conditioned Policy}
We show that our model can generate risk-bounded policies starting at random positions in a large maze, as visualized in Fig.~\ref{fig:flytrapbig}. When there exist obstacles between the start and the goal (i.e. when starting at the lower left or upper right room), our model outputs a risk-bounded path that balances between plan quality and safety.


\section{Conclusions}
In conclusion, we propose a risk conditioned soft actor critic model that generates risk-bounded policies in the context of chance-constrained MDPs.
Our model leverages a risk critic to estimate the execution risk at a given state and acting according to the policy, and adds a penalty term to the policy loss if the estimated risk is greater than the risk bound to produce risk-bounded policies. By providing upper risk bound as part of the input, our method is able to generate plans with different risk-averse levels for the agent on the fly. We demonstrate the advantage of our model in terms of both time complexity and path quality compared to a MILP-based baseline in a simple maze environment, and further validate its performance with more complex agent dynamics and larger state space, which are usually hard to handle by MILP-based methods.
Future work includes validating our model in more complicated settings (i.e. a larger state space, a variety of geometries, real-world experiments) and extending our algorithm to handle dynamic obstacles.

\section{Acknowledgment}
We thank Cheng Fang for the implementations of the IRA algorithm and comments on related work. We gratefully acknowledge funding support in part by the Toyota Research Institute grant 6944668.

\bibliographystyle{IEEEtran}
\bibliography{references} 

\end{document}

%% file: notations.tex
\def\BibTeX{{\rm B\kern-.05em{\sc i\kern-.025em b}\kern-.08em
    T\kern-.1667em\lower.7ex\hbox{E}\kern-.125emX}}

\newcommand{\E}[2]{\operatorname{\mathbb{E}}_{#1}\left[#2\right]}

\newcommand{\state}{\mathbf{s}}

\newcommand{\st}{{\state_t}}

\newcommand{\stp}{{\state_{t+1}}}

\newcommand{\action}{\mathbf{a}}

\newcommand{\at}{{\action_t}}

\newcommand{\Q}{Q}

\newcommand{\policy}{\pi}

\newcommand{\params}{\theta}

\newcommand{\pparams}{{\phi}}   

